\documentclass{article}

\usepackage{arxiv}

\usepackage[utf8]{inputenc} 
\usepackage[T1]{fontenc}    
\usepackage{hyperref}       
\usepackage{url}            
\usepackage{booktabs}       
\usepackage{amsfonts}       
\usepackage{nicefrac}       
\usepackage{microtype}      
\usepackage{lipsum}		

\usepackage{amssymb}
\usepackage{todonotes}
\usepackage{float}
\usepackage{tabularx}
\usepackage{booktabs}
\usepackage{multirow}
\usepackage{makecell}
\usepackage{utfsym} 
\usepackage{url}
\usepackage{amsmath}

\usepackage{threeparttable}
\usepackage{algorithm}
\usepackage{algorithmic}

\usepackage{doi}
\newcommand{\chk}{\usym{1F5F8}}

\title{Smart Parking with Pixel-Wise ROI Selection for Vehicle Detection Using YOLOv8, YOLOv9, YOLOv10, and YOLOv11}

\author{
    Gustavo P. C. P. da Luz \\
    Institute of Computing\\
    University of Campinas (UNICAMP)\\
    Av. Albert Einstein, 1251, \\
    Campinas, 13083-852, SP, Brazil \\
    \texttt{ra271582@students.ic.unicamp.br} \\
    \And
    Gabriel Massuyoshi Sato \\
    Institute of Computing\\
    University of Campinas (UNICAMP)\\
    Av. Albert Einstein, 1251, \\
    Campinas, 13083-852, SP, Brazil \\
    \texttt{ra172278@students.ic.unicamp.br} \\
    \And
    Luis Fernando Gomez Gonzalez\thanks{Corresponding Author}\\
    Institute of Computing\\
    University of Campinas (UNICAMP)\\
    Av. Albert Einstein, 1251, \\
    Campinas, 13083-852, SP, Brazil \\
    \texttt{gonzalez@unicamp.br} \\
    \And
    Juliana Freitag Borin\\
    Institute of Computing\\
    University of Campinas (UNICAMP)\\
    Av. Albert Einstein, 1251, \\
    Campinas, 13083-852, SP, Brazil \\
    \texttt{juliana@ic.unicamp.br} \\
}

\renewcommand{\shorttitle}{Smart Parking with Pixel-Wise ROI Selection for Vehicle Detection Using YOLO}

\hypersetup{
pdftitle={A template for the arxiv style},
pdfsubject={q-bio.NC, q-bio.QM},
pdfauthor={David S.~Hippocampus, Elias D.~Striatum},
pdfkeywords={First keyword, Second keyword, More},
}

\begin{document}
\maketitle

\begin{abstract}
	The increasing urbanization and the growing number of vehicles in cities have underscored the need for efficient parking management systems. Traditional smart parking solutions often rely on sensors or cameras for occupancy detection, each with its limitations. Recent advancements in deep learning have introduced new YOLO models (YOLOv8, YOLOv9, YOLOv10, and YOLOv11), but these models have not been extensively evaluated in the context of smart parking systems, particularly when combined with Region of Interest (ROI) selection for object detection. Existing methods still rely on fixed polygonal ROI selections or simple pixel-based modifications, which limit flexibility and precision. This work introduces a novel approach that integrates Internet of Things, Edge Computing, and Deep Learning concepts, by using the latest YOLO models for vehicle detection. By exploring both edge and cloud computing, it was found that inference times on edge devices ranged from 1 to 92 seconds, depending on the hardware and model version. Additionally, a new pixel-wise post-processing ROI selection method is proposed for accurately identifying regions of interest to count vehicles in parking lot images. The proposed system achieved 99.68\% balanced accuracy on a custom dataset of 3,484 images, offering a cost-effective smart parking solution that ensures precise vehicle detection while preserving data privacy.
 
\end{abstract}

\keywords{Smart Parking \and IoT \and Edge Computing \and YOLO}

\section{Introduction}
\label{sec:intro}
With the growing number of vehicles in cities, the need for improved parking management in public areas has become more pressing. The search for available parking spaces, commonly known as "cruising," contributes to increased traffic circulation on roads, leading to higher congestion and carbon emissions from vehicles~\cite{SAHARAN2020622}. Additionally, it causes driver dissatisfaction, which can potentially reduce economic competitiveness in areas where finding parking is difficult~\cite{Weinberger2020}. Motivated by the concept of smart cities, which aim to optimize resource and energy use and enhance service efficiency~\cite{8623018}, there is a strong drive to improve this process.

On average, 31\% of the land in large cities is used for parked cars, with some cities like Los Angeles~\cite{ASurveyofSmartParkingSolutions}, reaching up to 81\%. Additionally, with rapid urban population growth, the UN estimates that around 6 billion people will be living in cities by 2050~\cite{8623018}. Given these figures, studies related to smart cities, including this one, are essential for improving the efficiency of urban spaces in the future.

Several existing smart parking solutions use sensors in each parking space to detect occupancy. Others employ car sensors, monitoring cameras or even drones to capture parking lot images \cite{SmartParkingUnicamp}. Despite some successful implementations, many of these solutions have limitations. Scalability is a key factor, given the difficulty of deploying a complex Internet of Things (IoT) infrastructure for data collection and analysis in urban areas. Implementation costs are also a decisive factor in choosing the best solution for an ongoing project~\cite{8623018}. Privacy concerns regarding data and image collection from citizens are additional challenges.

This paper proposes an efficient and scalable smart parking solution by integrating IoT, Edge Computing, and Deep Learning. It explores the use of recent YOLO models for vehicle detection, addressing the limitations of existing methods. The paper evaluates different models using a pixel-wise Region of Interest (ROI) selection method, which improves flexibility and precision in detecting vehicles within parking images. One key scalability metric considered is the cost per parking space, with the goal of reducing costs while maintaining accurate predictions of the number of parked cars.

The proposed smart parking system uses cameras to capture images of the parking area at predefined intervals, according to the business needs. These images are processed by a neural network to determine the number of available parking spaces, which can be done either on a local edge device or on a remote server in the cloud. The number of vehicles is then sent to an IoT platform, enabling users to access this data through a website or mobile app before arriving at the parking facility. A real parking lot is used as a case study. The main contributions of this paper are: 

\begin{itemize}
    \item proposal of a cost-effective and scalable system for vehicle detection, demonstrating a high accuracy on a custom dataset using recently released pre-trained models and optimized image processing techniques;
    \item comparison of Edge Computing and Cloud Computing for image classification using neural networks, focusing on inference time across six different devices. This comparison highlights the efficiency of Edge Computing in scenarios where latency and accuracy are critical; 
    \item introduction of a new, fully customized masking method for ROI selection in images. This method offers greater flexibility and precision compared to traditional approaches by allowing free-form selection rather than just polygons, making it well-suited for complex scenarios;
    \item to the best of the authors' knowledge, this is the first paper to compare recently released models such as YOLOv9, YOLOv10, and YOLOv11 applied to a parking lot dataset;
    
    
    \item comparison of inference times for four different YOLO versions across six different hardware platforms with varying computational capacities. Additionally, we measured the latency in a standardized GPU environment (NVIDIA A100), which is not found on the literature. 
    
    
\end{itemize}

The rest of this paper is organized as follows: Section \ref{sec:relwork} brings an overview of the related works and similar approaches using camera-based systems. Section \ref{sec:sysmodel} presents an overview of the system and the processing scenarios. Section \ref{sec:experiments} discusses the experiments conducted and the metrics used. Section \ref{sec:results} presents the results and discusses model performance, time, resources, and cost analysis. Section \ref{sec:conclusion} concludes the paper and suggests possible directions for future work.

\section{Related Work}
\label{sec:relwork}
Currently, there are various methods for detecting cars in parking spaces, with the predominant method being the use of devices based on infrared, ultrasonic, or magnetic sensors. 

Infrared sensors can be divided into two types: i) passive detection, which detects changes in ambient infrared radiation when a car occupies a space - this type of sensor is also used in security alarms and automatic lighting systems; ii) active detection, which emits infrared radiation and measures the reflected signal to detect objects. This approach is also used in obstacle detection applications.
Ultrasonic sensors emit sound waves, and detection occurs based on the time it takes for the wave to return. The use of magnetometers for car detection is based on changes in the magnetic field when a car approaches the sensor.

While sensor usage proves to be efficient in producing accurate results, it has implementation and scalability limitations. The need for one or more sensors for each parking space poses a resource usage challenge, especially for large areas, significantly raising project costs~\cite{amato2017deep, Channamallu2023}. Additionally, a robust infrastructure is required for each sensor to work correctly, making the solution less scalable for large parking lots. Another limitation is that this solution is often applied in controlled indoor environments (e.g., underground mall parking). In open areas with significant pedestrian traffic, sensors may struggle to distinguish between cars and people, animals, or even cars not parked in a space~\cite{Xie2021}.

Another car detection strategy involves distributing tags and sensors for each vehicle~\cite{pala2007smart}. This method allows radiofrequency receivers to identify the number of cars in the parking lot. However, determining the exact location of vacant spaces is not tipically feasible, and the installation process can be complex~\cite{idris2009car}. 

Finally, car identification through images is a method for assessing the status of a parking lot. Drones can take aerial photos, but recognizing cars in vertical images poses a significant challenge~\cite{Moranduzzo2014}. Furthermore, issues related to power consumption and flight time add to the complexity of utilizing drones for this purpose~\cite{Srivastava2021}. The use of cameras mounted on poles is a promising method for capturing images from suitable angles for car recognition.


Before the advancements in machine learning and deep learning, classification was performed using methods like the traditional Haar Feature-based Cascade Classifier. This method achieved a detection rate ranging from 89.3\% to 95\% on a custom dataset, as reported by Sieck et al. [2020]\cite{sieck2020machine}. The study used an NVIDIA Jetson Nano for inference.

Acharya et al. [2018]\cite{acharya2018real} proposed the use of a classifier combining a Convolutional Neural Network (CNN) with a binary Support Vector Machine (SVM), achieving 99.7\% accuracy for the PKLot dataset~\cite{de2015pklot} and 96.7\% for a custom dataset. 

Bura et al. [2018]\cite{Bura2018} proposed a custom-designed neural network model based on AlexNet to detect parking occupancy from video streamed from top-view cameras in the dataset from the NVIDIA AI City Challenge
\cite{naphade2017nvidia}, combined with data from PKlot and CNRPark. The proposed network consists of 1 input layer, 1 convolution layer, 1 Rectified Linear Unit (ReLU), 1 max pooling, and 3 fully connected layers. It was designed to run on an edge devices such as the NVIDIA Jetson Tx2 or a Raspberry Pi. The proposed system also includes a set of cameras installed close to the ground to capture images of license plates for vehicle tracking. The authors achieved high precision in parking spot detection, with a  99.7\% success rate on a custom dataset combined with PKlot and CNRPark~\cite{amato2017deep}. The inference time per parking spot was 7.11~milliseconds. 

Carrasco et al. [2021]\cite{carrasco2021t} proposes T-YOLO, a modification of the YOLOv5 model tailored for tiny object detection. This approach achieves a precision of up to 96,34\% on the PKLot dataset with a fine-tuned model. It employs an ROI mask to select the monitored area following the pre-processing method described in Section \ref{subsec:preproc}.

Shukla et al. [2022]\cite{Shukla2022} presents a system that utilizes surveillance cameras installed in parking spaces to capture frames. A CNN model then identifies free or occupied slots. The system provides real-time updates to a server, which can be accessed through mobile and web applications. Additionally, it incorporates a Long Short-Term Memory (LSTM) model to predict parking space availability based on factors such as day, time, and weather conditions. The authors report that the CNN model achieves an precision of around 97.89\% under different weather conditions using the CNRPark dataset and around 97.5\% in a custom dataset. The proposed system uses a centralized architecture. 

Nithya et al. [2022]\cite{Nithya2022} uses Faster Region-based CNN (Faster R-CNN) with YOLOv3 to detect vehicles in parking lot images captured by a web camera in a NVIDIA Jetson Tx2. The accuracy results for this approach reach 98.41\% on a custom dataset. 

In 2023, Satyanath et al.\cite{Satyanath2023} attack the problem of vacant parking slot detection under hazy conditions. The detection system utilizes two networks: an end-to-end dehazing network and a parking slot classifier. For the first one, they employ an All-in-One Dehazing network (AOD-Net) architecture, while for the second one, they follow the mAlexnet architecture (proposed by Amato et al. [2016]\cite{Amato2016}). The proposed strategy improved the precision of the model proposed by Amato et al. [2016]\cite{Amato2016} by 10\% to 15\%. Runtime analysis did not consider edge devices.
Rafique et al. [2023]\cite{rafique2023optimized} uses a fine-tuned YOLOv5 to implement a parking management system, evaluating the model for PKLot dataset and custom dataset with respective accuracies of 99.5\% and 96.8\%, using the pre-processing ROI method showed in Section \ref{subsec:preproc}. The authors found that using a pre-trained YOLO model achieved results
superior to training with the PKLot dataset.

Recents advances can be seen in works such as the one from Doshi et al. [2024]\cite{doshi2024comparision}, that compared the performance of YOLOv3, YOLOv5, YOLOv7, YOLOv8 across PKlot and an Aerial View Car Detection dataset with custom training, achieving the best results with YOLOv8. Sundaresan Geetha et al. [2024]\cite{sundaresan2024comparative} evaluated how well YOLOv8 and YOLOv10 could detect different vehicles, showing a higher classification accuracy of cars for YOLOv8. 

Hudda et al. [2024]\cite{hudda2024wsn} used R-CNN and Faster R-CNN combined with multiple wireless sensor networks (WSN) in a smart parking, optimizing energy usage and switching between WSN and vision based sensing. They achieved 99.16\% accuracy with a fine-tuned Faster R-CNN with a ResNet50 backbone at the ACPDS dataset \cite{marek2021image}. Despite using WSN and optical verification at the edge, the work does not perform the inference on-device, thus it was not considered as edge inference.

Considering all types of car recognition solutions in parking lots, the study conducted in this paper addresses many challenges faced by these systems, such as using only pre-trained state of the art models. Using cameras for car detection eliminates the need for a robust and expensive hardware structure, as in the case of sensors. Additionally, the project is implemented in an open parking area with considerable foot traffic, prepared to handle these challenges. Data privacy will also be possible to be maintained, as image processing is possible to happen at the edge of the system, eliminating the need to transmit camera pictures over the internet to a cloud or server.

The definition of the edge can vary across different authors. In this context, the concept of Edge Intelligence (EI) is adopted, which represents a fusion of Edge Computing and Artificial Intelligence. This work follows the definition of Zhou et al. [2019] \cite{zhou2019edge}, who proposed a six-level rating system for EI based on the amount and path length of data offloading. In this case, a Level 3 EI device is considered, where a deep neural network model is trained in the cloud, but inference is performed on the device itself. This approach ensures that only the number of vehicles detected is sent to the cloud, preserving data privacy. While using vision-based approaches for vehicle detection can present challenges in terms of data privacy, it also offers an opportunity to leverage EI to reduce data transmission costs \cite{plastiras2018edge}.

A device is considered to be at the edge of the network when it performs inference locally on the device, rather than sending the collected data to a central location, such as a data center hosted by a cloud service provider, or any other external server or computer that requires data transfer over the network.

Table~\ref{tab:parking_detection_comparison} compares different approaches to the proposed solution in terms of inference location (edge/cloud), the use of a mask for selecting a ROI, and the classifier model employed.

\begin{table}[H]
    \centering
    \setlength{\tabcolsep}{4pt}
    \renewcommand{\arraystretch}{2} 
    \caption{Comparison of Camera-based Parking Occupancy Detection Methods}
    \label{tab:parking_detection_comparison}
    \begin{tabular}{l|cc|c|c}
        \hline
        \multirow{2}{*}{Work} & \multicolumn{2}{c|}{Inference at} & \multirow{2}{*}{\makecell{ROI \\ Mask}} & \multirow{2}{*}{Classifier} \\
        \cline{2-3}
        & Cloud & Edge & & \\
        \hline
        Acharya et al.\cite{acharya2018real} (2018) & \chk & & & CNN + SVM \\
        Bura et al.\cite{Bura2018} (2018) &  & \chk & & Custom AlexNet \\
        Sieck et al.\cite{sieck2020machine}(2020)  & & \chk & & Haar Cascade \\
        Carrasco et al.\cite{carrasco2021t} (2021) & \chk &  & \chk & Custom YOLOv5 \\
        Shukla et al.\cite{Shukla2022} (2022) & \chk & & & CNN + LSTM \\
        Nithya et al.\cite{Nithya2022} (2022) &  &\chk &  & \makecell{Faster R-CNN \\ and YOLOv3} \\
        Satyanath et al.\cite{Satyanath2023} (2023) & \chk & & & \makecell{AOD-Net \\ and mAlexNet} \\
        Rafique et al.\cite{rafique2023optimized} (2023) & \chk & & \chk & YOLOv5 \\
        Hudda et al.\cite{hudda2024wsn} (2024) & \chk & & & \makecell{Faster R-CNN \\ and R-CNN} \\
        Doshi et al.\cite{doshi2024comparision} (2024) & \chk & & & \makecell{YOLO: \\ v3, v5, v7 and v8} \\
        \hline
        \textbf{This work} (2024) & \chk & \chk & \chk & \makecell{YOLO: \\v8, v9, v10 and v11 }\\
        \hline
    \end{tabular}
\end{table}

\section{System model}
\label{sec:sysmodel}
\begin{figure}[H]
\centering
\includegraphics[width=12.5 cm]{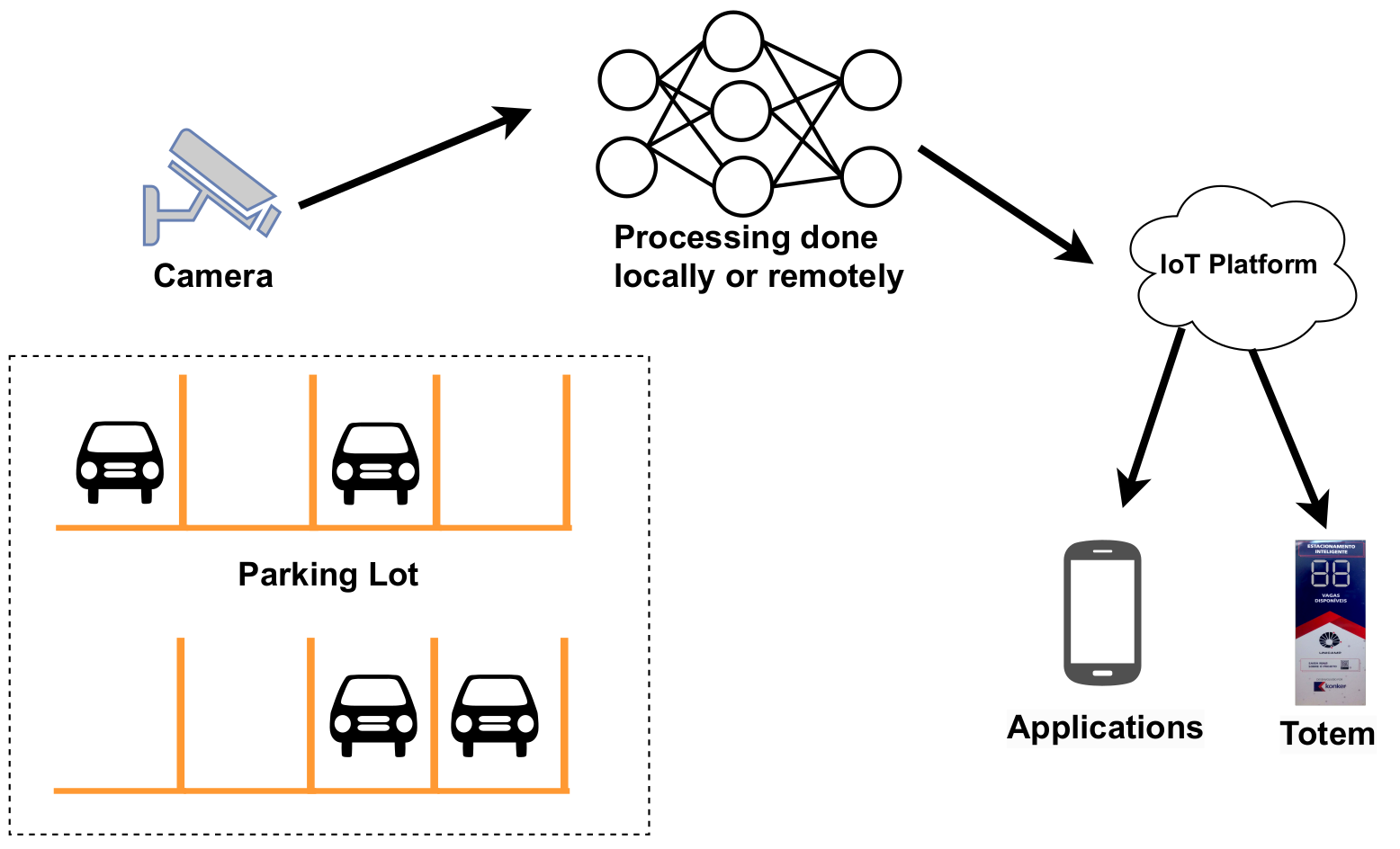}
\caption{System Architecture.\label{layout}}
\end{figure}

The main idea of the system is to use a camera to monitor several parking spaces by capturing images at predefined time intervals. Each image goes through a deep neural network model trained to detect objects. This step can be performed on a server in the cloud by sending the image through the Internet or on a local device, leveraging the potential of edge computing. We tested both approaches in this work. Next, the result of the inference is used to count the number of available spots, deducting the number of detected vehicles from the total number of spots. This result is sent to an IoT platform, which maintains a history of data and makes the data available for applications. Figure \ref{layout} shows the layout of the system.

There are significant differences in processing power, bandwidth usage, privacy, and cost when choosing the location for image processing. Taking that into account when designing a data pipeline is a necessary step in the system model proposal, choosing which data is captured, what is the processing at the edge and how the cloud deals with this data \cite{munn2020staying}. This can provide an architecture that benefits from edge and cloud while meeting the application business needs.  



\subsection{Using a cloud server to process}

In this scenario, images taken from the parking lot are sent to a cloud server. The advantage of this approach is the reduced processing time, as the computing power in the cloud is superior to that of a device at the edge. However, there are disadvantages such as increased bandwidth usage for sending images, bureaucratic hurdles, and privacy issues associated with transmitting images from public places.



\subsection{Using an edge device to process}

In this scenario, a device such as a Raspberry Pi is connected to the cameras to provide edge processing. This alternative avoids the transmission of images over the network, reducing bandwidth usage and enhancing privacy. However, it presents challenges such as inference time, requiring a model that can detect cars with high accuracy but low computational cost, given the limited hardware resources of edge devices compared to those available on cloud servers. Specifically at the smart parking application, we want the device that senses the data (eg. a Raspberry Pi) to be the device that makes an inference of the number of vehicles detected, sending only this number to the cloud. 

\section{Experiments}
\label{sec:experiments}
To conduct the experiments, the dataset was collected, labeled, pre-processed, and filtered. The inference model options were chosen and the number of cars predicted in each image was compared with the labels to obtain performance metrics for the proposed model. The experiments were conducted using various hardware configurations to compare both cloud and edge computing strategies.

\subsection{Dataset}

\begin{figure}[H]
\centering
\includegraphics[width=10.5 cm]{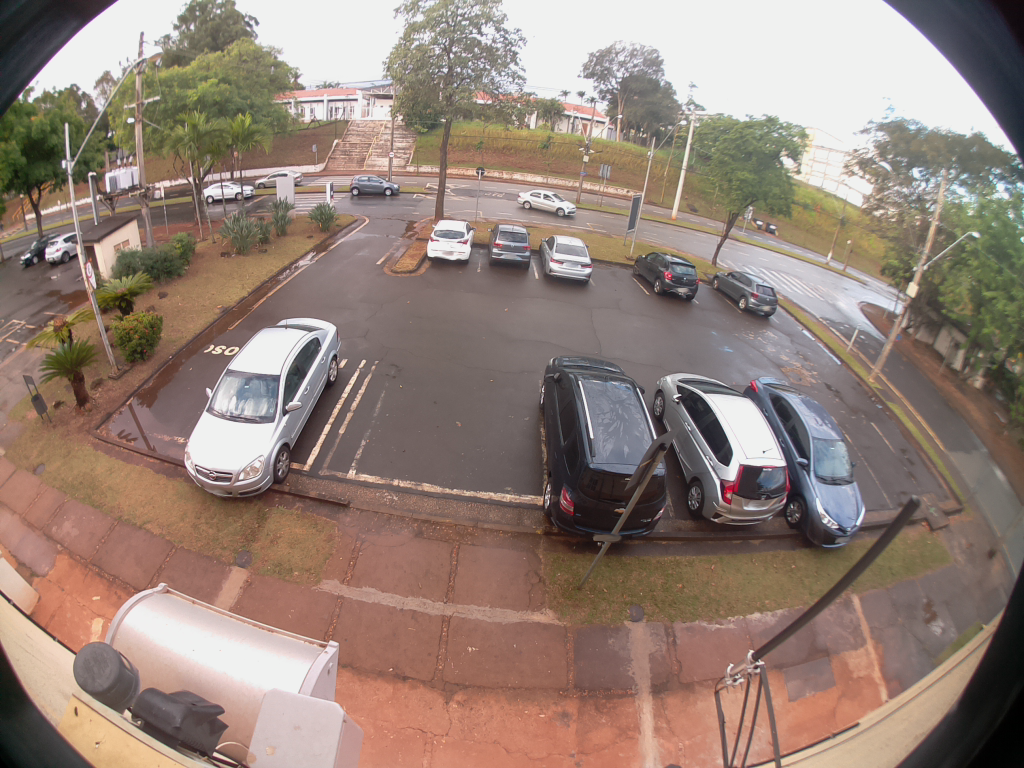}
\caption{A sample picture of the dataset.\label{samplepic}}
\end{figure}

In a parking lot within Universidade Estadual de Campinas (UNICAMP), we installed a camera. A single photo was sufficient to cover the entire parking lot, which contained 16 spaces, as depicted in Figure \ref{samplepic}. In parking lots with favorable viewing angles, dozens of spaces can be monitored using just one camera. A prior study was conducted to determine the optimal viewing angle for camera placement.


\textit{Data collection:} For several weeks, photos were captured to compile and validate our dataset. The entire dataset contains 4,477 images selected to classify 71,632 different parking scenarios, including rainy, sunny, cloudy, day, night, crowded, and empty conditions.

\textit{Data labeling:} We labeled the dataset, counting the number of vehicles in each image, including cars and trucks. This number of vehicles is deducted from the total number of places to obtain the amount of free spaces.

\textit{Data pre-processing:} Data was pre-processed by compressing the images from Portable Network Graphics (PNG) format to Joint Photographic Experts Group (JPG).


\textit{Data filtering:} The dataset was filtered before running the experiments to images containing at least one car, so images, where the parking is empty, are not counted. This helps to balance the number of backgrounds and vehicles and produce fair metrics. 22\% of the original dataset was removed, resulting in a number of 3,484 images.

\subsection{Inference Model}

A lot of classifiers could be used and compared, as, there is a wide range of possible models in terms of object detection, such as shown by Table \ref{tab:parking_detection_comparison}. Sapkota et al. [2024]\cite{sapkota2024yolov10} made an analysis of the advancements from traditional approaches such as classic machine learning classifiers, followed by the emergence of CNN's, Region-based CNN, and the YOLO series. As stated by other works superior results of YOLO \cite{sumit2020object}\cite{sato2024tvbox} and initial tests not showing promising results with models such as Mask R-CNN and EfficientDet, we selected different recent flavors of YOLO in this study, with versions 8, 9, 10, and 11, selecting the largest and lightest model of each version (eg. YOLOv8x and YOLOv8n).

YOLO stands for "You Only Look Once", being an
object detector that uses features learned by a deep convolutional neural network to detect an object in a single forward pass, different than two-stage detectors that propose regions and then process the region candidates \cite{tamang2023enhancing}. For YOLO, every input image to the model is split into grids and each cell in a grid predicts some bounding boxes and gives a confidence value which indicates how sure the
model is that the box contains an object. After this step, the model knows where the object is in that image but it does not know what that object is. For knowing the class, each cell predicts a class probability using the pre-trained weights. Finally, the box and class predictions are combined to identify the object. 

The first version of YOLO was released in 2015 by Redmon [2016]\cite{redmon2016you} with an CNN combined with two fully connected layers. The evolution is shown by Wang and Liao [2024]\cite{wang2024yolov1}, highlighting the changes in the architectures, such as the backbone, usage of Feature Pyramid Networks (FPN) ~\cite{lin2017feature}, Spatial Pyramid Pooling Networks (SPP) ~\cite{he2015spatial} and Path Aggregation Networks (PAN) ~\cite{liu2018path}. This work focuses on the latest versions released since 2023: YOLOv8, YOLOv9, YOLOv10, and YOLOv11.


YOLOv8~\cite{jocher2023ultralytics} uses an anchor-free model with a decoupled head, sigmoid function for activation, and softmax function for class probabilities, using the backbone as a variant of CSPNet with the named cross-stage partial bottleneck with two convolutions (C2f) module, combining high-level features with contextual information to improve performance \cite{terven2023comprehensive}. After going through the CNN, the FPN is replaced by a PAN.

YOLOv9~\cite{wang2024yolov9} implements Programmable Gradient Information to preserve data across the layers of neural network and improve convergence, combined with a change in the backbone proposing the use of GELAN, combining CSPNet and ELAN. 

YOLOv10~\cite{wang2024yolov10} on the other hand brings attention mechanisms with the position-sensitive attention block (PSA) to the CSPNet backbone, while eliminating the need for Non Maximum Suppression (NMS) during inference. 

YOLOv11~\cite{jocher2024ultralytics} brings improvements in the feature extractor using more convolutions with a modified CSPnet backbone, containing a series of PSA blocks enhancing attention mechanisms and further data augmentations techniques.

Some metrics of the detection on the benchmark Common Objects in Context (COCO) dataset \cite{lin2014microsoft} of 2017 with 80 pre-trained classes and diverse objects are shown in Table \ref{tabresultsyolococo}.


\begin{table}[H]
\caption{Comparison of Model Performance Metrics for COCO dataset. Adapted from Jocher et al. [2023]~\cite{jocher2023ultralytics}.}
\label{tabresultsyolococo}
\begin{threeparttable} 
\begin{tabularx}{\textwidth}{XXXXXX}
\toprule
\small\textbf{Model} & \small\textbf{mAP\tnote{1}}   & \small\textbf{Reported Latency\tnote{2} (ms)} & 
\small\textbf{Measured Latency\tnote{3} (ms)} & 
\small\textbf{Parameters\tnote{4} (M)} & \small\textbf{FLOPs\tnote{5} (B)}\\
\midrule
YOLOv8n & 37.3  & 6.16 & 3.61 $\pm$ 0.38 & 3.2 & 8.7 \\
YOLOv8x & 53.9  & 16.86 & 8.42 $\pm$ 0.49 & 68.2 & 257.8 \\
YOLOv9t & 38.3  & - & 5.12 $\pm$ 0.41 & \textbf{2.0} & 7.7 \\
YOLOv9e & \textbf{55.6}  & - & 10.69 $\pm$ 0.39 & 58.1 & 192.5 \\
YOLOv10n & 38.5  & 1.84 & \textbf{3.14 $\pm$ 0.11} &2.3 & 6.7 \\
YOLOv10x & 54.4  & 10.7 & 7.46 $\pm$ 0.28 & 29.5 & 160.4 \\
YOLOv11n & 39.5  & \textbf{1.5} & 3.61 $\pm$ 0.21 & 2.6 & \textbf{6.5} \\
YOLOv11x & 54.7  & 11.3 & 7.87 $\pm$ 0.42 & 56.9 & 194.9 \\
\bottomrule
\end{tabularx}
\begin{tablenotes}
\footnotesize 
    \item[1] mean Average Precision (mAP) measures object detection performance across multiple classes over different Intersection over Union (IoU) thresholds, from 0.5 to 0.95.
    \item[2] Latency is measured on different GPU's running TensorRT.
    \item[3] Latency is measured on a NVIDIA A100 GPU running TensorRT.
    \item[4] Parameters refer to the number of trainable weights in the model.
    \item[5] Floating Point Operations (FLOPs) measure the computational complexity of the model.
\end{tablenotes}
\end{threeparttable} 
\end{table}


The analysis of the model performance metrics for the COCO dataset suggests that YOLOv9e has the highest detection accuracy, with a mAP of 55.6. YOLOv11n, in the other hand, is more efficient, with the lowest reported latency and computational complexity. This shows that more advanced version of YOLO are getting more adequate for resource-constrained devices, with little impact on accuracy. As shown in Table \ref{tabresultsyolococo}, some values are missing in the reported latency, and the measurements are not consistent. In some cases, an A100 GPU is used (e.g., for YOLOv8), while in others, a T4 GPU is used (e.g., for YOLOv10 and YOLOv11), and in some instances, no latency data is reported (e.g., for YOLOv9). This inconsistency motivated the introduction of a standardized latency measure using only an NVIDIA A100 GPU with TensorRT models at the FP32 precision. The test involved executing 1,500 inferences on a COCO dataset image, discarding the initial 100 inferences. The results from this test show that YOLOv10 achieved lower latency. Section \ref{sec:results} presents additional inference time measurements across six devices, providing a more comprehensive assessment of the latest models' latency, including those in non-GPU environments.


The dataset was not used to train the model, as we used only the pre-trained weights of the network, so 100\% of the data was used as validation. In our work, we kept the premise of feeding the model with the images at original collection sizes (768x1024), as the model performs an automatic resizing in the image if they do not match the original training shape (640x640), when using the Ultralytics library \cite{jocher2023ultralytics}. Vehicles were counted based on the result generated by the model inference, matching the class names stored in the model dictionary.

\subsection{Region Of Interest}

\begin{figure}[H]
\centering
\includegraphics[width=10.5 cm]{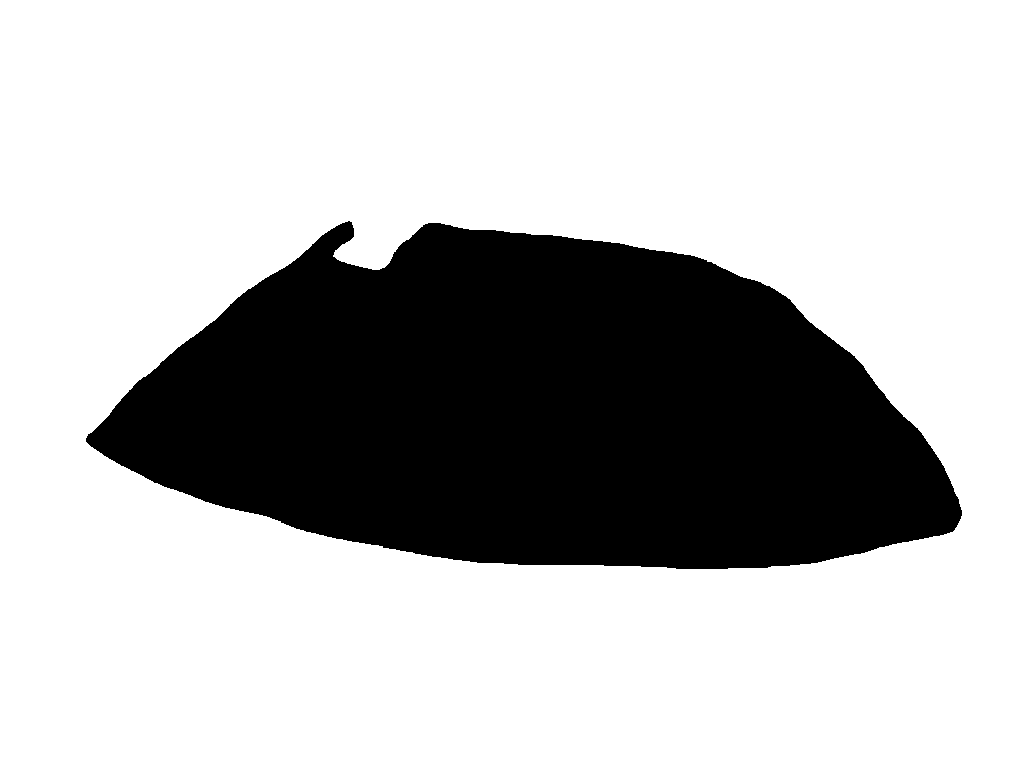}
\caption{Reference Mask used to select ROI .\label{maskpic}}
\end{figure}

The dataset contains images with vehicles positioned outside of the parking lot space, so we created a mask to select the ROI (Region Of Interest) avoiding classifying cars in unwanted places.  We used two approaches to select the ROI: with a pre-processing of the photos and with post-processing after the neural network makes the prediction. For both approaches, the same mask was created by selecting a random picture and manually editing it using the free image editor software GNU Image Manipulation Program (GIMP) \footnote{\url{https://www.gimp.org}}. The region in which there are wanted cars was selected and painted with black color, leaving the rest as white, as shown in Figure \ref{maskpic}. The mask was left in the same shape as the original image. 

\subsubsection{Pre-Processing Mask Approach}
\label{subsec:preproc}

\begin{figure}[H]
\centering
\includegraphics[width=10.5 cm]{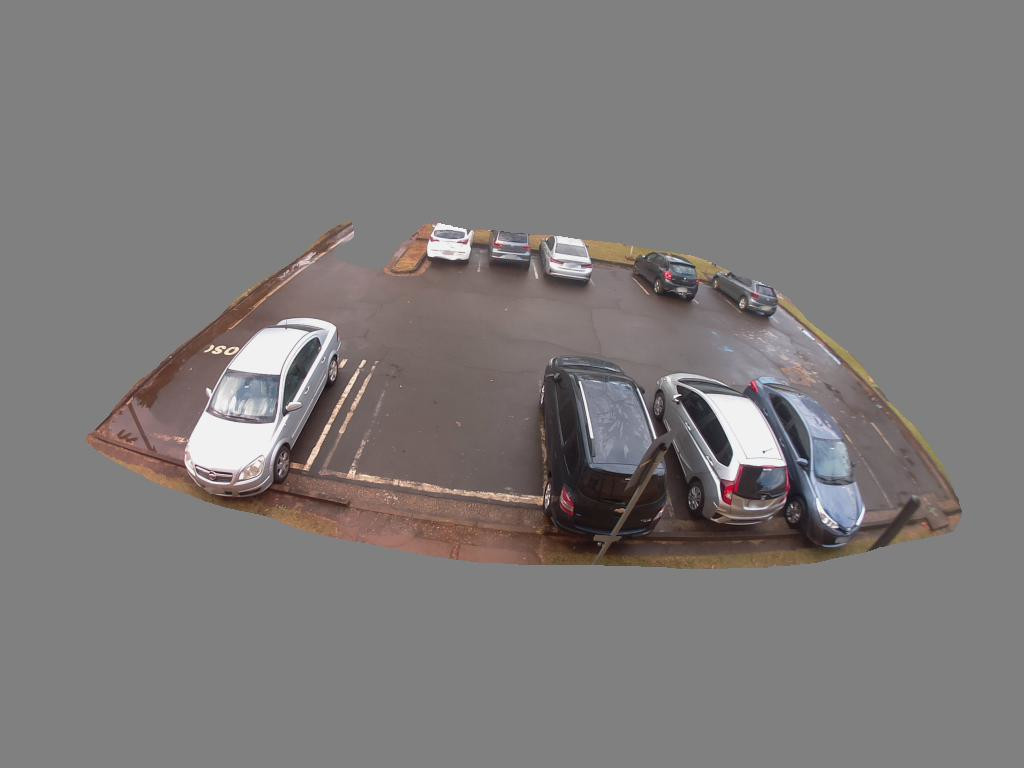}
\caption{Sample of the pre-processing approach.\label{maskpreproc}}
\end{figure}

In the pre-processing approach, we utilized the reference mask to change the value of all pixels corresponding to the white region to gray in the three color channels (R = 128, G = 128, B= 128). This effectively covered the regions outside of the parking lot with gray, as can be seen in Figure \ref{maskpreproc}. Works \cite{carrasco2021t,rafique2023optimized} have used similar handcrafted masks in order to select the region and to consider only annotated slots when fine-tuning YOLOv5 models, fixing issues at missannotated datasets. All the images underwent this process before being passed to the model. Algorithm \ref{alg:preprocessing_mask} shows the pseudocode, where each image pixels are changed to contain only the selected region: 

\begin{algorithm}[H]
\caption{Pseudocode of Pre-Processing Mask Approach}
\label{alg:preprocessing_mask}
\begin{algorithmic}[1]
\STATE \textbf{Input:} Load reference mask (black for ROI, white for non-ROI)
\FOR{each image in the dataset}
    \STATE Load image
    \STATE Multiply image by mask
    \FOR{each pixel in the image}
        \IF{pixel in reference mask is white}
            \STATE Set pixel value in the image to gray in the three color channels (128, 128, 128)
        \ENDIF
    \ENDFOR
    \STATE Run model inference to detect objects in the updated image
    \STATE \textbf{Output:} Total object count for the image
\ENDFOR
\end{algorithmic}
\end{algorithm}

\subsubsection{Post-Processing Mask Approach}

The pre-processing approach modifies the image pixels, removing parts that can be important to provide context to the model, which is a factor that can improve the performance of YOLO models \cite{zhao2024cmca}\cite{oreski2023yolo}\cite{krois2021impact}.

To solve that, the proposed post-processing approach uses the same reference mask shown at Figure \ref{maskpic} and do not change any pixel value of all the images inputted into the model. As shown in Algorithm \ref{alg:postprocessing_mask}, after the model produces the results containing the bounding boxes of the predictions, we compare the center of x and y coordinates with the corresponding pixel value at the reference mask and considers only the vehicles inside it, which had the black color on the mask. 

\begin{algorithm}[H]
\caption{Pseudocode of Post-Processing Mask Approach}
\label{alg:postprocessing_mask}
\begin{algorithmic}[1]
\STATE \textbf{Input:} Load reference mask (black for ROI, white for non-ROI)
\FOR{each image in the dataset}
    \STATE Load image
    \STATE Run model inference to detect objects in the image
    \STATE \textbf{Initialize:} object\_count
    \FOR{each predicted bounding box representing the object class}
        \STATE Get the center coordinates $(x, y)$ of the bounding box
        \IF{pixel value at $(x, y)$ in reference mask is black}
            \STATE Increment object\_count by 1 (valid ROI)
        \ENDIF
    \ENDFOR
    \STATE \textbf{Output:} Total object count for the image
\ENDFOR
\end{algorithmic}
\end{algorithm}

\begin{figure}[H]
\centering
\includegraphics[width=10.5 cm]{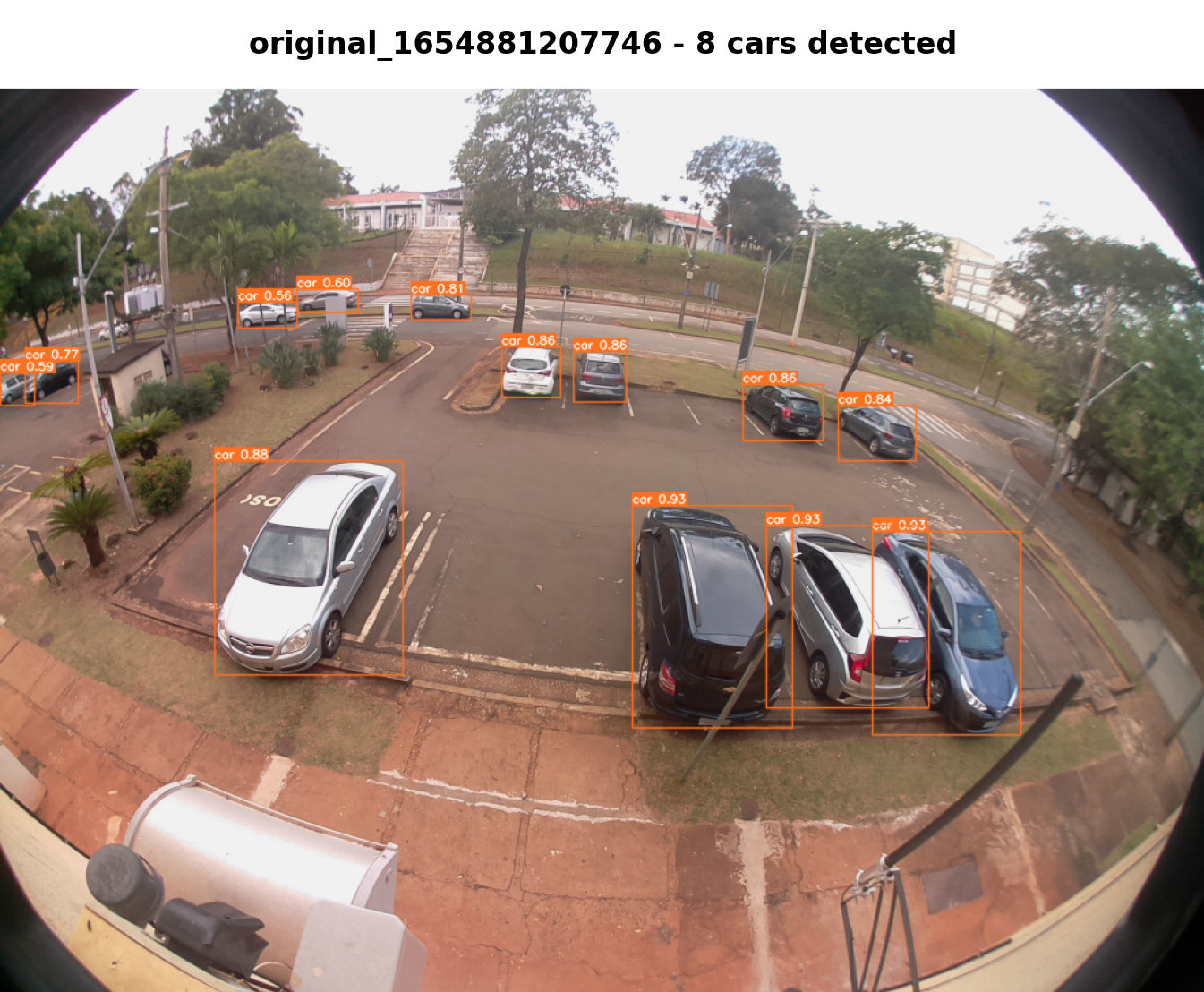}
\caption{Sample of the post-processing approach.\label{maskpostproc}}
\end{figure}

In Figure \ref{maskpostproc} it is possible to see that there are 13 cars identified, however only 8 are counted, which is the desired effect computing only vehicles inside the parking lot. 

Although this post-processing method represents a novel approach for selecting a fully customized pixel-wise ROI for object detection, similar methods can be found in the Ultralytics library \cite{jocher2023ultralytics}, particularly in the "Object Counting in Different Regions" section and other tutorials. However, these existing methods are limited to regular polygonal shapes and may not perform well in more complex, custom scenarios. In contrast, the proposed method is both precise and flexible, accommodating any mask shape.

\subsection{Metrics}

To compute the metrics and compare the results between different approaches and with existing works, the following basic terms are defined:

\begin{itemize}
    \item True Positives (TP): Correctly predicted empty space.
    \item True Negatives (TN): Correctly predicted vehicle.
    \item False Positives (FP): Predicted empty space, but there's a vehicle.
    \item False Negatives (FN): Predicted vehicle, but there's an empty space.
\end{itemize}

This choice was based on similar smart parking solutions ways of assessing model performance \cite{Bura2018,polprasert2019camera}. The following metrics were used based on the terms defined:

\begin{enumerate}
    \item Accuracy provides an overall view of the model's performance. It is calculated as:
    \begin{equation}
    \text{Accuracy} = \frac{TP + TN}{TP + FP + TN + FN}
    \end{equation}

    \item $F_1$-score is the harmonic mean of precision and recall, especially useful for imbalanced classes. 
    \begin{itemize}
        \item Precision measures the proportion of true positive predictions among all positive predictions and is calculated as:
        \begin{equation}
        \text{Precision} = \frac{TP}{TP + FP}
        \end{equation}

        \item Recall measures the proportion of true positive predictions among all actual positive instances and is calculated as:
        \begin{equation}
        \text{Recall} = \frac{TP}{TP + FN}
        \end{equation}
    \end{itemize}
    The $F_1$-score is defined as:
    \begin{equation}
    \text{$F_1$-score} = 2 \times \frac{\text{precision} \times \text{recall}}{\text{precision} + \text{recall}}
    \end{equation}

    \item Balanced Accuracy is useful for imbalanced datasets.
    \begin{itemize}
        \item Sensitivity measures the proportion of true positive predictions among all actual positive instances and is calculated as:
        \begin{equation}
        \text{Sensitivity} = \frac{TP}{TP + FN}
        \end{equation}

        \item Specificity measures the proportion of true negative predictions among all actual negative instances and is calculated as:
        \begin{equation}
        \text{Specificity} = \frac{TN}{TN + FP}
        \end{equation}
    \end{itemize}
    Balanced Accuracy is defined as:
    \begin{equation}
    \text{Balanced Accuracy} = \frac{\text{Sensitivity} + \text{Specificity}}{2}
    \end{equation}
\end{enumerate}


We also produced the confusion matrices for both pre and pos-processing approaches, which is a matrix that is often used to describe the performance of a classification model on a set of test data for which the true values are known.

\section{Results and Discussion}
\label{sec:results}

\subsection{Model Performance}


\begin{table}[H] 
\caption{Model Performance Metrics: $n = 3484$.\label{tabresultsmodelperformance2}}
\begin{tabularx}{\textwidth}{XXXXXXXX}
\toprule
\textbf{ROI Method} & \multicolumn{4}{c}{\textbf{Model Metrics}} \\
\cmidrule(lr){2-5}
& \textbf{Model} & \textbf{Accuracy} & \textbf{Bal Acc} & \textbf{F$_1$-score} \\
\midrule
\multirow{2}{*}{\shortstack[l]{Pre-Processed\\ Mask}}
&  YOLOv8n & 82.21\% & 66.24\% & 0.9172 \\
&  YOLOv9t & 84.08\% & 59.09\% & 0.8950 \\
&  YOLOv10n & 89.73\% & 70.20\% & 0.9300 \\
&  YOLOv11n & 85.75\% & 61.23\% & 0.9066 \\
\midrule
&  YOLOv8x & 93.47\% & 76.57\%  & 0.9579 \\

&  YOLOv9e & 94.26\% & 79.45\% & 0.9604\\

&  YOLOv10x & \textbf{95.04\%} & \textbf{80.37\%} & \textbf{0.9674} \\

&  YOLOv11x & 94.30\% & 78.99\%  & 0.9623 \\

\midrule
\midrule
\multirow{2}{*}{\shortstack[l]{Post-Processed \\Mask}} 
&  YOLOv8n & 96.63\% & 94.86\% & 0.9720 \\
&  YOLOv9t & 94.60\% & 86.90\% & 0.9644 \\
&  YOLOv10n & 97.57\% & 97.32\% & 0.9757 \\
&  YOLOv11n & 96.15\% & 95.10\% & 0.9599 \\
\midrule
&  YOLOv8x & 98.88\% & 98.49\%  & 0.9907 \\

&  YOLOv9e & \textbf{99.76\% }& \textbf{99.68\%}  & \textbf{0.9975} \\

&  YOLOv10x & 98.75\% & 98.30\% & 0.9875 \\

&  YOLOv11x & 99.46\% & 99.39\% & 0.9946 \\

\bottomrule
\end{tabularx}
\end{table}

\begin{figure}[htbp]
\centering
\includegraphics[width=13.5 cm]{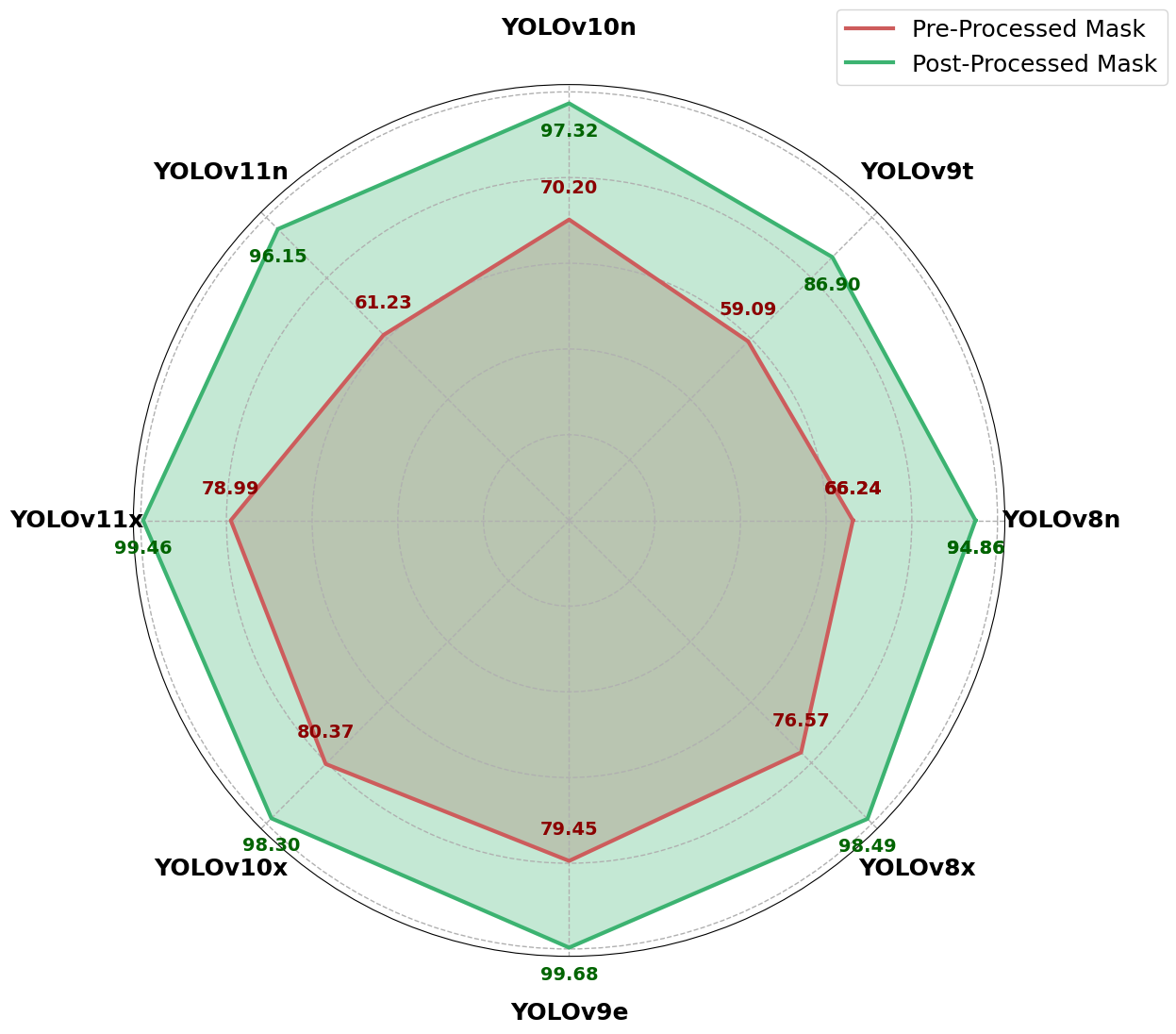}
\caption{Balanced Accuracy by model and ROI method.\label{figbalaccroi}}
\end{figure}

Table ~\ref{tabresultsmodelperformance2} shows the results for the selected metrics for both methods and selected deep learning models. To compare the models we use a radar chart that shows the balanced accuracy for each model and ROI method in Figure \ref{figbalaccroi}.

\begin{figure}[htpb]
\centering
\includegraphics[width=0.49\textwidth]{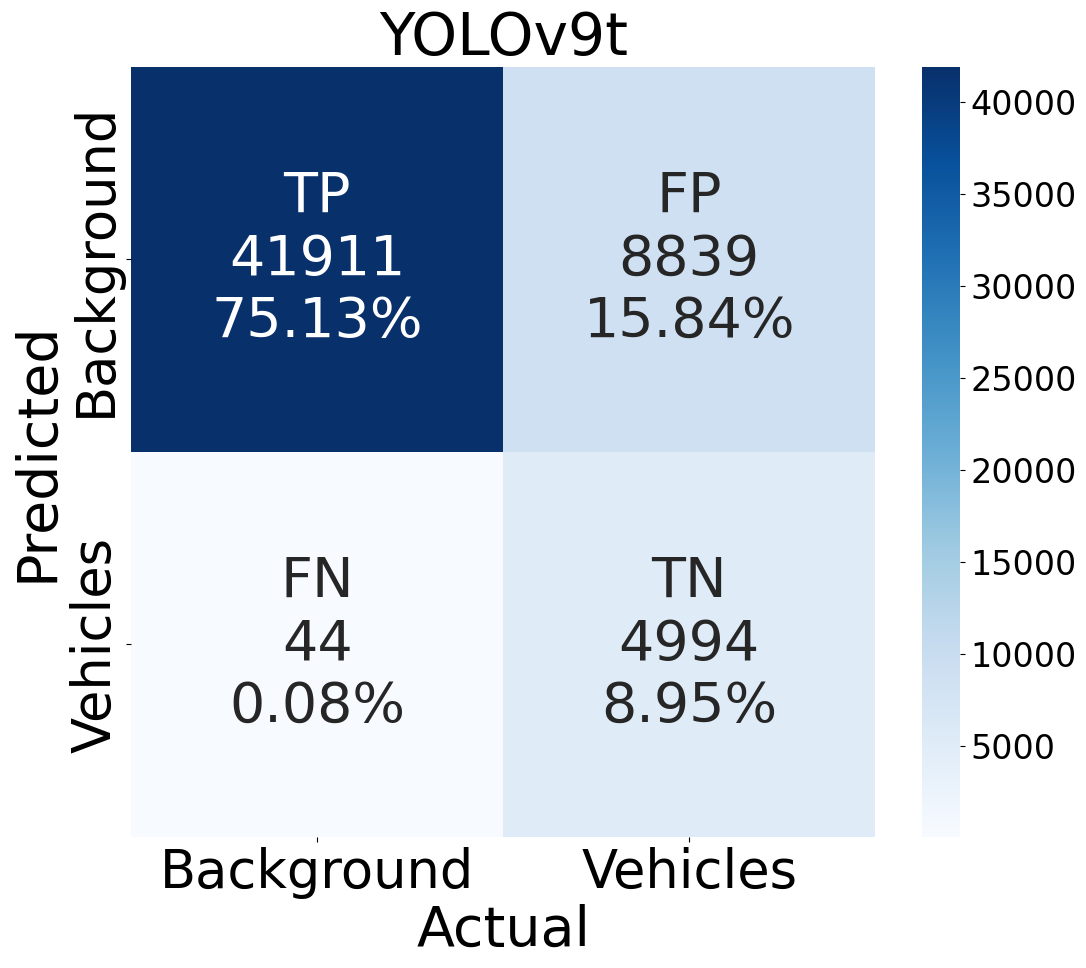}
\includegraphics[width=0.49\textwidth]{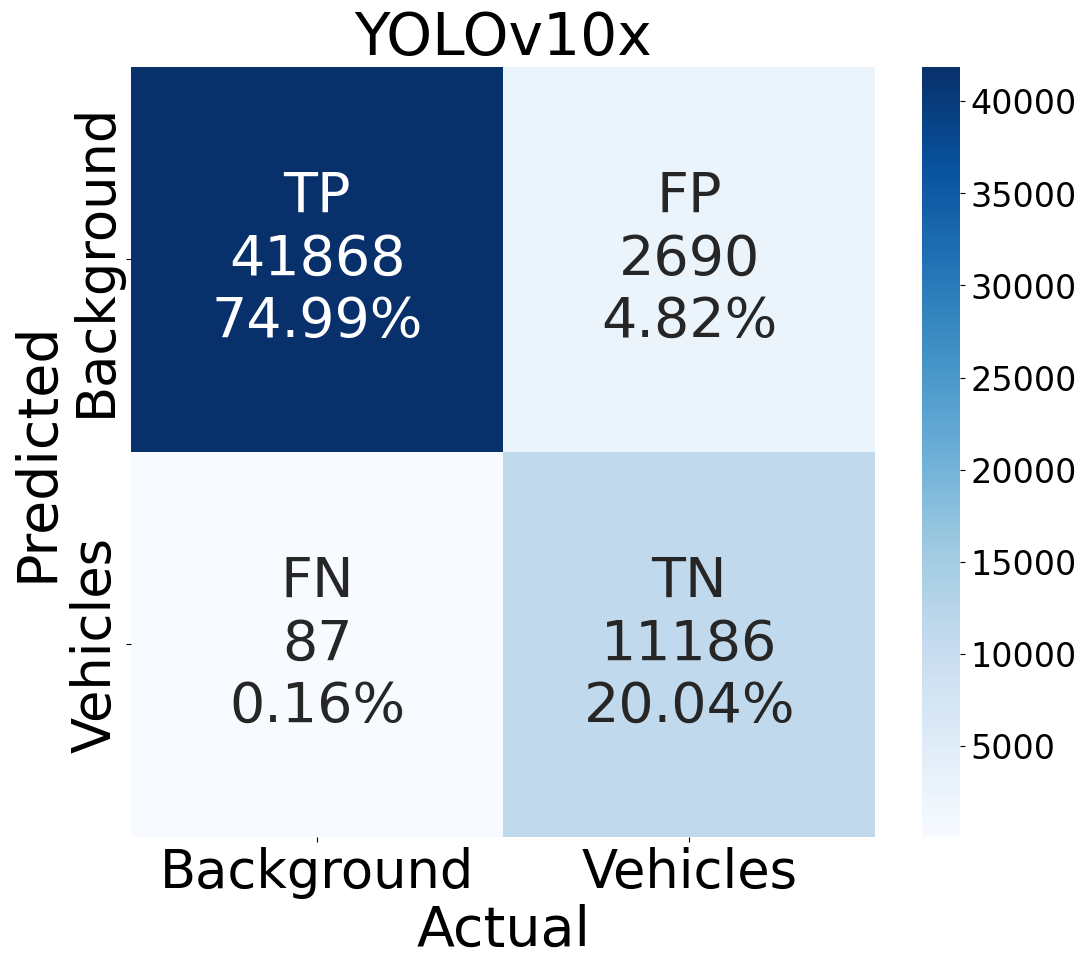}
\caption{Confusion Matrices for worst (YOLOv9t) to best (YOLOv10x) result on pre-processing method based on balanced accuracy.\label{figconfmatpre}}
\end{figure}

\begin{figure}[htpb]
\centering
\includegraphics[width=0.49\textwidth]{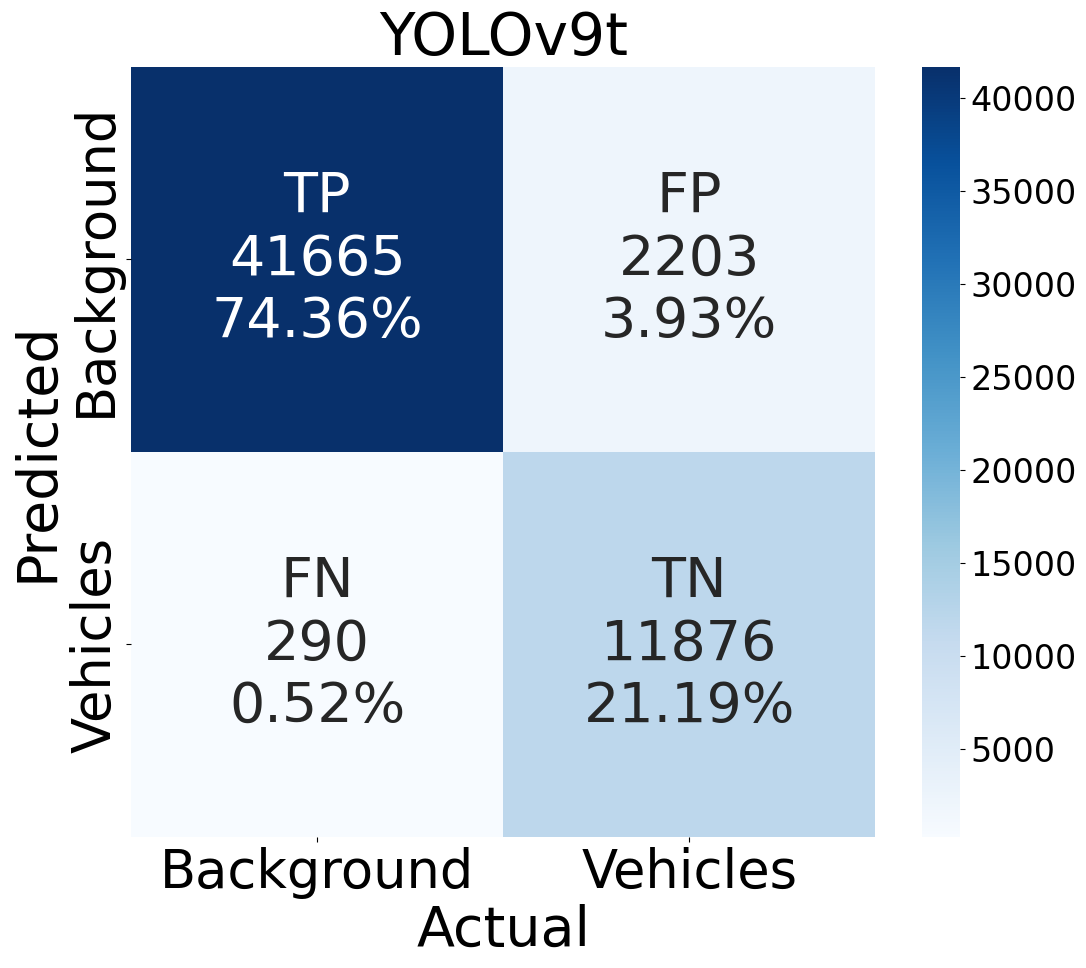}
\includegraphics[width=0.49\textwidth]{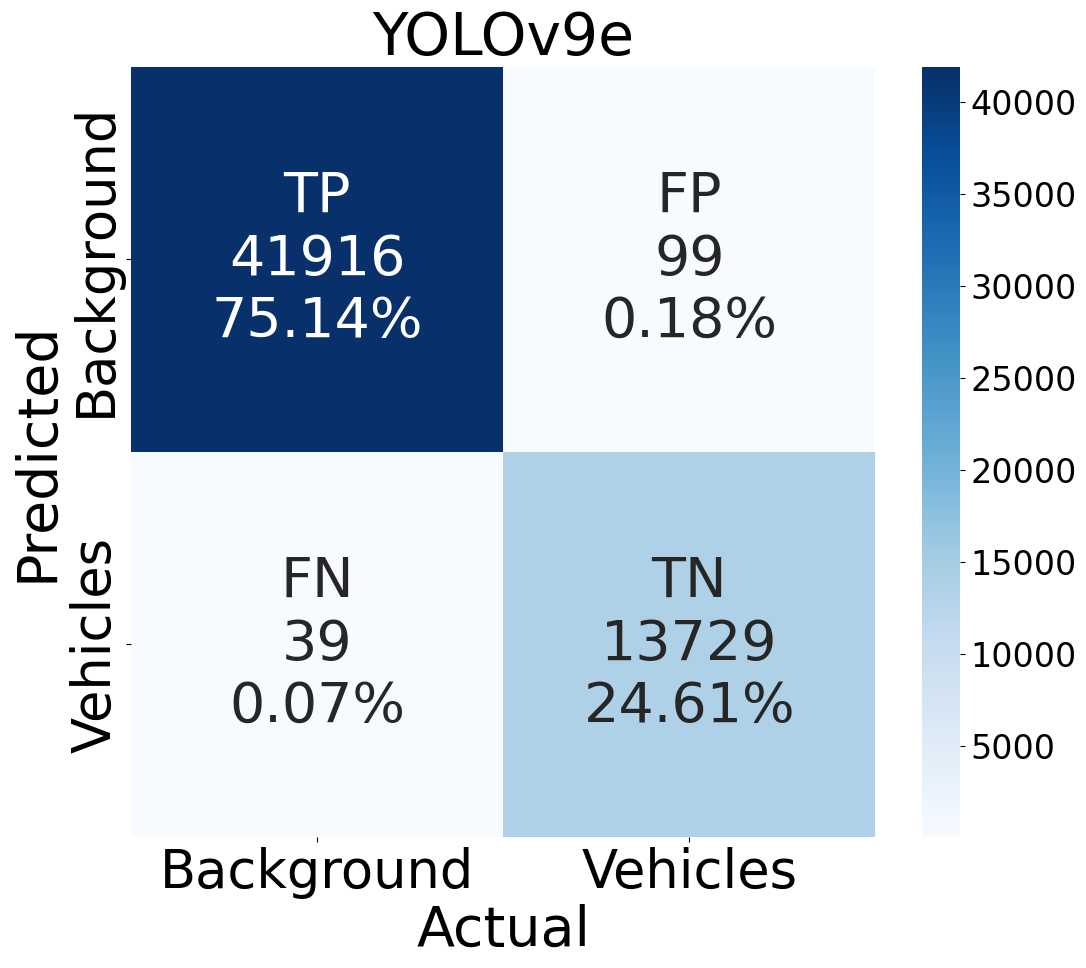}
\caption{Confusion Matrices for worst (YOLOv9t) to best (YOLOv9e) result on post-processing method.\label{figconfmatpost}}
\end{figure}

It is possible to observe that despite the model chosen, the post-processed method showed higher balanced accuracy, with significant gains such as over 28 percent points at YOLOv8n. It was possible to notice that, as expected, the lightest models performs worse than the heaviest, but the range of this variation is different according to the model evaluated. For example, specially when analysing YOLOv9t and YOLOv9e with the post-processed method, the balanced accuracy goes from 86.90\% to 99.68\%, indicating a range of around 13 percent points. This variation was not strongly observed at the other YOLO versions. YOLOv10, in particular, showed to be an stable model for both ROI methods, despite not having the highest performance. YOLOv11 presented a high performance specially at the extra large version, however still lower than YOLOv9, which was the better model in terms of accuracy, balanced accuracy and $F_1$-score. To continue the analysis, we investigated the confusion matrices for the worst and best possible result for each ROI method at Figures \ref{figconfmatpre} and \ref{figconfmatpost}.


Regarding model performance, the pre-processing method showed a high amount of false positives. A possible reason for that is the lack of context in the image that got some regions removed from it, leaving some vehicles unclassified. Another possible cause is the need to leave some vehicles on the edges of the image not fully appearing, as the masks were delimited in a way no cars on the outside appear, which would lead to unwanted vehicles being also classified. This issue was attenuated for the best case scenario with YOLOv10x, still leaving a considerable amount of 4.82\% false positives. In other hand, the pre-processing method showed a small amount of false negatives for the worst and best model, indicating that when a vehicle is detected, it has a high confidence and the prediction can be trusted.

For the post-processing method, it is possible to observe a smaller amount of false positives, with its worst case (YOLOV9t with 3.93\%)  lower than the best case of the to pre-processing method (YOLOv10x with 4.82\%). The worst case showed to have a considerable amount of false negatives (YOLOV9t with 0.52\%), higher than the pre-processing method, which can be caused by cases where the model detects more than one vehicle in a part of the image. However, for the best model found, it was the lowest number found for all cases (YOLOv9e with 0.07\%). This is an evidence that choosing the appropriate model and method can result in a well balanced confusion matrix detecting more vehicles and maintaining a high detection confidence.

\begin{figure}[]
    \centering
    \includegraphics[width=0.7\textwidth]{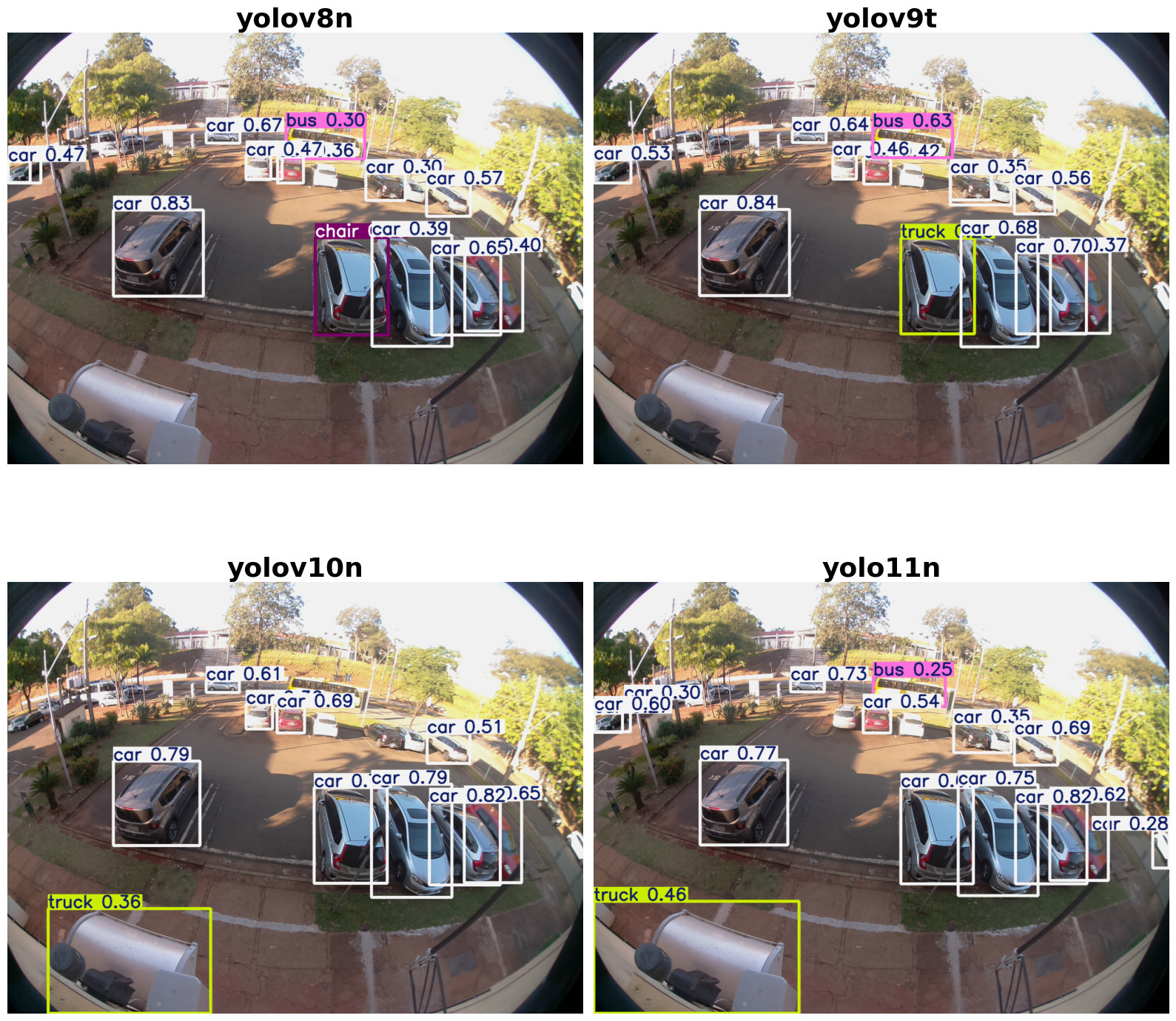}
    
    
    \includegraphics[width=0.7\textwidth]{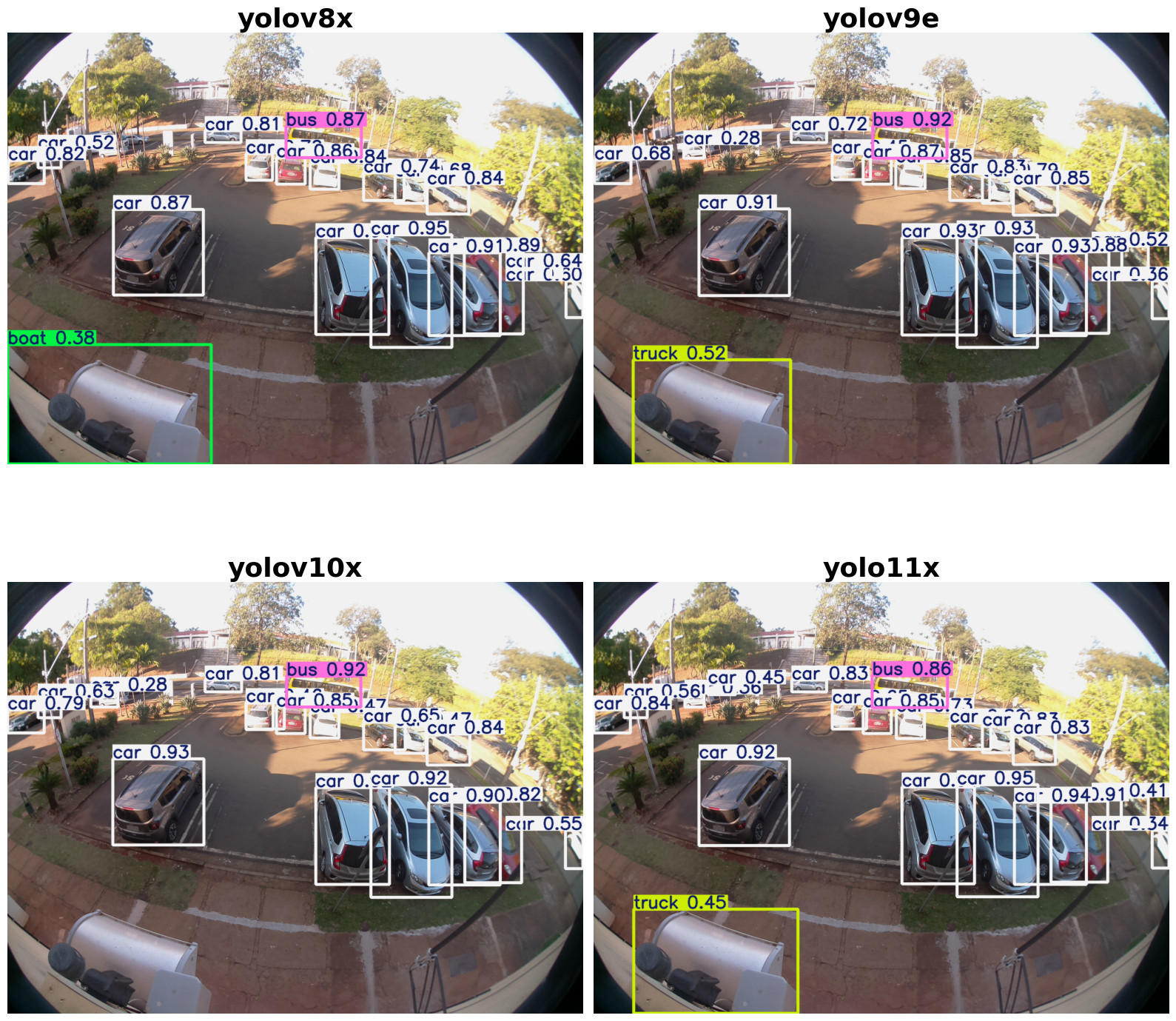}
    \caption{Qualitative analysis for the selected models}
    \label{fig:qualanalysis}
\end{figure}

We also made an qualitative analysis of the results of the post-processing method by selecting the tested models at Figure \ref{fig:qualanalysis}.

Smaller versions of the models suffer with lighting conditions and miss classify cars in some cases, presenting a larger number of false positives. In the largest versions, the predictions are more stable and the problems encountered are attenuated, despite in some cases having a higher number of false negatives.

\subsection{Time Performance}
\begin{table}[htbp]
\centering
\setlength{\tabcolsep}{11pt} 
\begin{threeparttable}
    \caption{Time Analysis Metrics}
    \label{tabresultstimeperformance}
    \begin{tabularx}{\textwidth}{l c c c c c c c c} 
        \toprule
        \textbf{Hardware} & \multicolumn{8}{c}{\textbf{YOLO Version}} \\
        \cmidrule(lr){2-9} 
        & \textbf{8n} & \textbf{8x} & \textbf{9t} & \textbf{9e} & \textbf{10n} & \textbf{10x} & \textbf{11n} & \textbf{11x} \\
        \midrule
        \makecell{Desktop \\ PC GPU\tnote{1}} & 
        \makecell{16 $\pm$ \\ 0.7 ms} & 
        \makecell{30 $\pm$ \\ 0.2 ms} & 
        \makecell{22 $\pm$ \\ 0.5 ms} & 
        \makecell{31 $\pm$ \\ 0.3 ms} & 
        \makecell{18 $\pm$ \\ 0.5 ms} & 
        \makecell{24 $\pm$ \\ 0.2 ms} & 
        \makecell{17 $\pm$ \\ 0.3 ms} & 
        \makecell{27 $\pm$ \\ 0.2 ms} \\
        \midrule
        \makecell{Mobile \\ PC GPU\tnote{2}} & 
        \makecell{26 $\pm$ \\ 0.5 ms} & 
        \makecell{204 $\pm$ \\ 0.7 ms} & 
        \makecell{29 $\pm$ \\ 0.6 ms} & 
        \makecell{198 $\pm$ \\ 0.9 ms} & 
        \makecell{29 $\pm$ \\ 0.6 ms} & 
        \makecell{161 $\pm$ \\ 1.2 ms} & 
        \makecell{27 $\pm$ \\ 1.1 ms} & 
        \makecell{164 $\pm$ \\ 2.0 ms} \\
        \midrule
        \makecell{Desktop \\ PC CPU\tnote{3}} & 
        \makecell{45 $\pm$ \\ 1.6 ms} & 
        \makecell{395 $\pm$ \\ 2.2 ms} & 
        \makecell{62 $\pm$ \\ 0.8 ms} & 
        \makecell{470 $\pm$ \\ 2.5 ms} & 
        \makecell{49 $\pm$ \\ 1.8 ms} & 
        \makecell{315 $\pm$ \\ 1.7 ms} & 
        \makecell{45 $\pm$ \\ 1.1 ms} & 
        \makecell{351 $\pm$ \\ 1.6 ms} \\
        \midrule
        \makecell{Mobile  \\PC CPU\tnote{4}} & 
        \makecell{73 $\pm$ \\ 1.7 ms} & 
        \makecell{953 $\pm$ \\ 3.3 ms} & 
        \makecell{102 $\pm$ \\ 8.7 ms} & 
        \makecell{968 $\pm$ \\ 4.5 ms} & 
        \makecell{87 $\pm$ \\ 2.3 ms} & 
        \makecell{728 $\pm$ \\ 3.4 ms} & 
        \makecell{80 $\pm$ \\ 3.0 ms} & 
        \makecell{801 $\pm$ \\ 2.5 ms} \\
        \midrule
        \midrule
        \makecell{Raspberry \\Pi 4} & 
        \makecell{1 $\pm$ \\ 0.02 s} & 
        \makecell{13 $\pm$ \\ 1.2 s} & 
        \makecell{1 $\pm$ \\ 0.01 s} & 
        \makecell{16 $\pm$ \\ 0.2 s} & 
        \makecell{1 $\pm$ \\ 0.01 s} & 
        \makecell{12 $\pm$ \\ 0.1 s} & 
        \makecell{1 $\pm$ \\ 0.03 s} & 
        \makecell{14 $\pm$ \\ 0.1 s} \\
        \midrule
        \makecell{Raspberry\\Pi 3} & 
        \makecell{2 $\pm$ \\ 0.06 s} & 
        \makecell{28 $\pm$ \\ 0.7 s} & 
        \makecell{2 $\pm$ \\ 0.02 s} & 
        \makecell{92 $\pm$ \\ 9.3 s} & 
        \makecell{2 $\pm$ \\ 0.03 s} & 
        \makecell{24 $\pm$ \\ 0.2 s} & 
        \makecell{2 $\pm$ \\ 0.02 s} & 
        \makecell{46 $\pm$ \\ 7.4 s} \\
        \bottomrule
    \end{tabularx}
    \begin{tablenotes}
    \footnotesize 
        \item[1] NVIDIA RTX4060Ti
        \item[2] NVIDIA MX450
        \item[3] AMD Ryzen 5 3600
        \item[4] Intel i7-11390H @ 3.40GHz
    \end{tablenotes}
\end{threeparttable}
\end{table}



\begin{figure}[btpb]
\centering
\includegraphics[width=12.5cm]{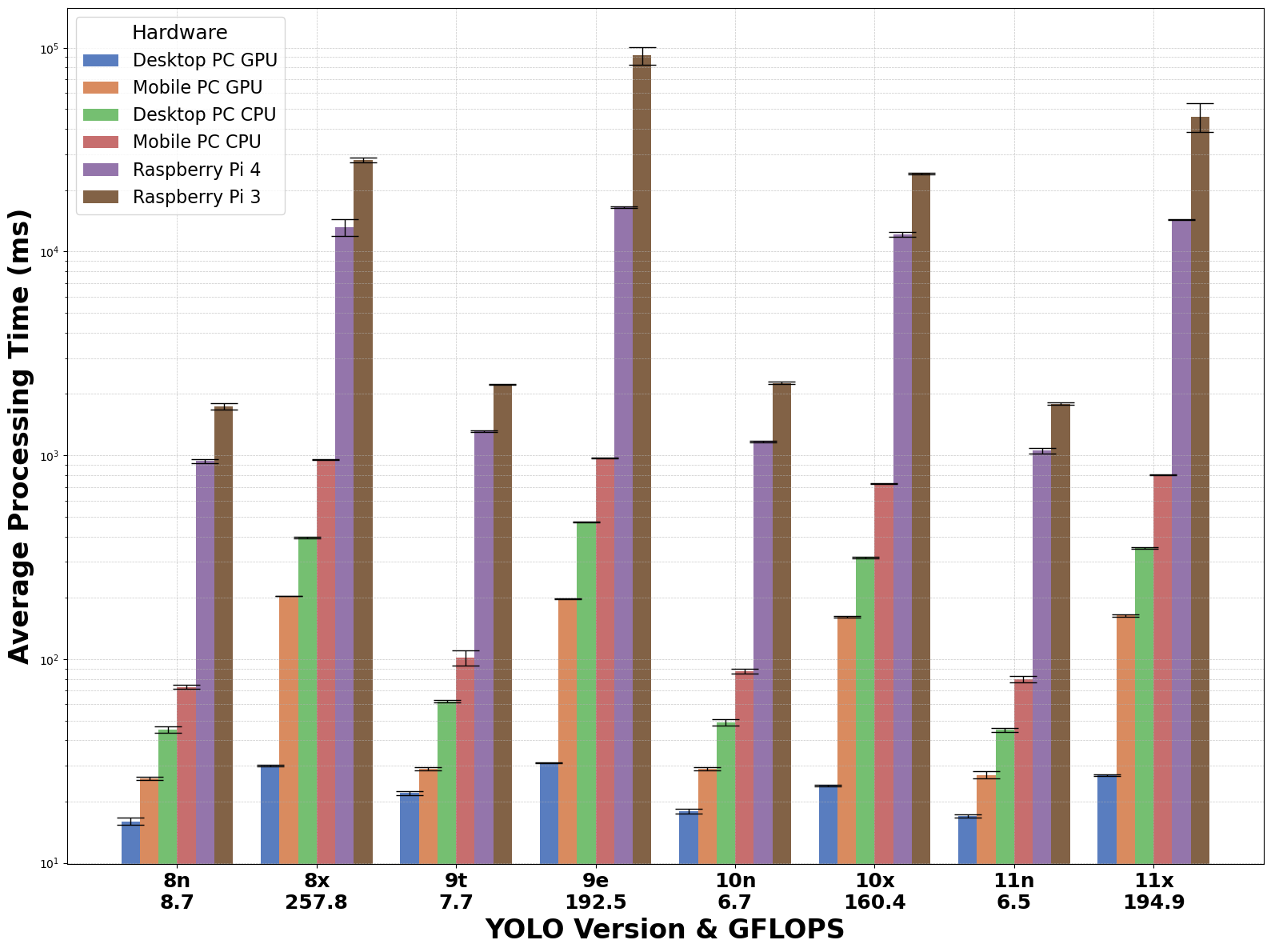}
\caption{Average Processing time for each hardware and model.\label{figavgtime}}
\end{figure}

Table ~\ref{tabresultstimeperformance} shows the results for the average time and standard deviation to perform model inference per picture in multiple hardware. The average time was taken by selecting 80 random pictures from the dataset and running model inference in each hardware. For all CPU tests, we used all threads available. These variations in inference time are a form of assessing the suitability of each device for applications, according to the business needs. In our case, a wait time up to five minutes is acceptable, since the parking lot usually do not suffer major differences in terms of occupation in this time window, however, other cases may need to choose more carefully the model and hardware setup. Figure \ref{figavgtime} shows inference time by model and hardware using the logarithm scale on the y-axis.

The Raspberry Pi units were kept in an environment with controlled temperature to prevent performance degradation due to CPU throttling. This was a crucial step for the stability observed in the results presented in the Table, as overheating caused variations in the time required for each image inference. Raspberry Pi 3 Model B+(1 GB) and Raspberry Pi 4 Model B (4 GB) were used to perform tests at edge devices, achieving an inference time range from 1-92 seconds, which is beyond acceptable for the application needs. 



The desktop computer is based on a Ryzen 5 3600 (6 cores / 12 threads) with 96 GB of DDR4 3200MHz RAM. The GPU used was the 16GB NVIDIA RTX 4060Ti. Similar to the Raspberry Pis, the PC was operated in an environment with forced ventilation and controlled temperature, and no performance degradation due to temperature increase was observed. Using the automatic method for selecting the GPU's performance states (P-states), it was observed that it did not consume more than 45W during the tests. This would be the hardware with most computational capabilities used at the study and a viable choice for cloud computing, achieving an almost real-time inference time range from 16 milliseconds in GPU to 470 milliseconds in CPU.

The mobile PC used for this test was a Dell Inspiron 15-5510 equipped with an Intel i7-11390H processor, with a NVIDIA MX450 GPU. One bottleneck observed during testing was CPU throttling, which affected inference times, underscoring the challenges of deploying high-demand models on such hardware, that presented an inference time range from three to five times higher than the GPU of the mobile PC. This reflects the current tendency at optimizing YOLO models for GPU. The Mobile GPU did not allow power consumption measurements as done on the desktop GPU; it was only possible to monitor the operating temperature, which remained constant at 76°C throughout the experiment.  


We previously investigated using YOLOv8x for object detection on Raspberry Pi 3 and Raspberry Pi 4 at a people counting application \cite{sato2024tvbox}, indicating that beyond being possible to use this hardware for edge processing in terms of inference time, average CPU and memory usage requirements are also met, even at more resource constrained devices such as TV Boxes \cite{da2024repurposing}. Table \ref{tabresultsyolococo} and \ref{tabresultstimeperformance} presents YOLOv8x as the model with highest FLOPs, parameters and latency. This indicates that optimized newest models should also be compatible with this devices without issues, as the state of the art advances. Newest nano version of YOLO have better latency on constrained resources devices, while extra large version have better latency on GPU. These results have significant implications for edge intelligence, as less resource constrained devices are developed and lighter more accurate deep learning models can be used near the sensor, maintaining data privacy.

\subsection{Cost Analysis}

We estimated the solution cost to be 177 United States Dollars (USD), composing a Raspberry Pi 4 Model B (4GB RAM) a surveillance camera (Raspberry Pi Camera Module 3), a power supply, a MicroSD Card and a weatherproof case. This estimate may vary slightly depending on the type of camera
used. As for the solution using sensors, which is currently the most used commercially, we calculate an average of 15 USD per parking space. To define this value we consider that a device can be used for only one parking spaces, and is composed mostly of a sensor, a microcontroller and a solar panel. The bill of materials for both kind of solutions are shown at Tables \ref{tab:camera_solution_cost} and \ref{tab:sensor_solution_cost}. 

\begin{table}[htbp]
    \centering
    \caption{Bill of Materials for the camera-based solution}
    \begin{tabular}{|l|c|c|}
        \hline
        \textbf{Item} & \textbf{Quantity} & \textbf{Cost (USD)} \\
        \hline
        Raspberry Pi 4 Model B (4GB RAM) & 1  & 55 \\
        Raspberry Pi Camera Module 3 & 1 &  25 \\
        Power Supply & 1  & 10 \\
        MicroSD Card & 1  & 15 \\
        Case (Weatherproof) & 1 & 15 \\
        \hline
        \textbf{Total} & 5 & \textbf{120} \\
        \hline
    \end{tabular}
    \label{tab:camera_solution_cost}
\end{table}

\begin{table}[htbp]
    \centering
    \caption{Bill of Materials for the sensor-based solution (assuming one device per space)}
    \begin{tabular}{|l|c|c|}
        \hline
        \textbf{Item} & \textbf{Quantity} & \textbf{Cost (USD)} \\
        \hline
        Microcontroller & 1  & 5 \\
        Small Solar Panel & 1  & 5 \\
        Battery & 1  & 6 \\
        Charge Controller & 1  & 2 \\
        Sensor & 1  & 2 \\
        Enclosure (Weatherproof) & 1  & 10 \\
        \hline
        \textbf{Total} & 6 & \textbf{30} \\
        \hline
    \end{tabular}
    \label{tab:sensor_solution_cost}
\end{table}

\begin{figure}[H]
\centering
\includegraphics[width=10.5 cm]{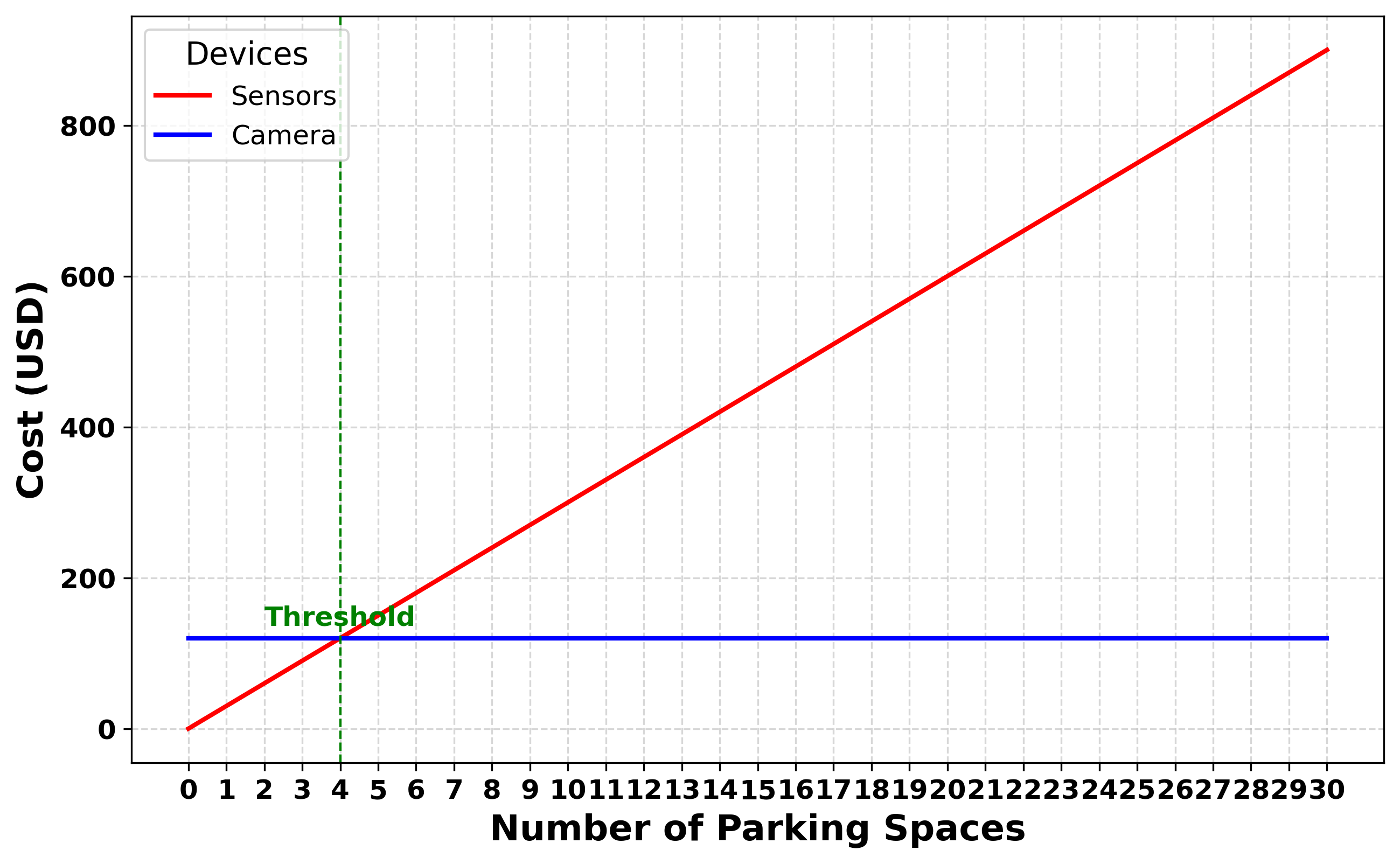}
\caption{Cost Comparison between Cameras and Sensors.\label{figcostcomp}}
\end{figure}

Figure \ref{figcostcomp} shows the estimated costs of the solutions using sensors and cameras, with the threshold where it pays off to use cameras is showed. 
Despite having a higher initial cost, the solution using cameras is more economical in larger parking lots since a single camera can monitor a higher number of parking spaces. In our case, as we count with 16 spots, we are at the 4x threshold point. Another aspect to consider is concerning the maintenance cost, as solutions using sensors have a high maintenance cost due to the high number of modules needed while our solution uses few modules that would eventually need maintenance, while camera-based solutions are easier and cheaper to monitor, as in some cases they are already implemented at the parking lot and can be reused.



\section{Conclusion}
\label{sec:conclusion}

This paper evaluates vehicle classification in various parking lot scenarios using a low-cost architecture comprised of accessible hardware and modern inference methods. This approach maintains data privacy by utilizing edge devices, such as a Raspberry Pi, achieving inference times ranging from 1 to 92 seconds. Future work could explore different methods to reduce model complexity, including quantization, conversion to TensorFlow Lite, and model pruning.

Overall, the proposed post-processing method to select the ROI in the image performed better than the traditional pre-processing method, solving the issue of vehicles left unclassified. The tested models presented good performance metrics, achieving a balanced accuracy of 99.68\% with YOLOv9e, which indicates that not always newest released YOLO models are having better results in terms of accuracy, but can bring deep learning closer to the edge with lightest models. This performance could be also compared with other variants of models and also with the regular polygonal ROI selection in future work.

As new models are released, that could be increased even more using other neural networks as object detection is a highly emerging field of research. An analysis can be made to understand how the models are performing according to different scenarios and other datasets can be used for testing. At bigger parking lots, other issues can appear as they contain a more diverse range of scenarios, with more vehicles and potentially deal with multiple cameras. This may indicate the need to use different techniques such as other image processing tools and fine-tuned models to achieve good performance. 

Using deep learning yields good results; however, it provides less insight into the model's features. The use of explainable AI can bring more trustworthiness and transparency to the users of smart city systems \cite{javed2023survey}. That can be combined with the different masks approaches to understand the impact on the neural networks using handcrafted masks. 

Another direction for future work involves conducting stability tests to assess how consistent the predictions are for similar images, as well as improving the labeling of the dataset by following bounding box coordinates. This would enable the calculation of additional metrics commonly used in image classification problems, such as Intersection over Union (IoU). Also, assessing inference time at different devices such as NVIDIA Jetson Nano, Raspberry Pi 5 and even TV Boxes could enable a benchmark with a larger range of edge devices options.

\section*{Declaration of competing interest}
The authors declare that they have no known competing financial interests or personal relationships that could have appeared
to influence the work reported in this paper.

\section*{Acknowledgments}
This project was supported by CAPES (process 88887.999360/2024-00), CNPq (process 308840/2020-8 and 131653/2023-7), by the Brazilian Ministry of Science, Technology and Innovations, with resources from Law nº 8,248, of October 23, 1991, within the scope of PPI-SOFTEX, coordinated by Softex and published Arquitetura Cognitiva (Phase 3), DOU 01245.003479/2024 -10, and by FAPESP (process 2023/00811-0).

\section*{Contributions}
All authors contributed to interpreting the results and writing and reviewing the manuscript. G.P.C.P.D.L., G.M.S., and L.F.G.G. conducted the experiments. J.F.B. supervised the project.

\section*{Availability of data and materials}
The datasets generated and/or analyzed during the current study are not available due to the sensitive nature of the research, however codes and materials that support the findings of this study are available upon reasonable request. All license plates that could appear in this work are already unidentified at the base picture and also were blurred.

\section*{Abbreviations}{
The following abbreviations are used in this paper:\\

\noindent
\begin{tabular}{@{}ll}
AI & Artificial intelligence \\
AOD-Net & All-in-One Dehazing Network\\
ARM & Advanced RISC Machine\\
CNN & Convolutional Neural Network\\
COCO & Common Objects in Context\\
CPU & Central Processing Unit\\
C2f & Cross-Stage Partial Bottleneck with Two Convolutions\\
EI    & Edge Intelligence (EI) \\
Faster R-CNN & Faster Region-based Convolutional Neural Networks\\
FLOPs & Floating Point Operations \\
FN & False Negatives\\
FP & False Positives\\
GIMP & GNU Image Manipulation Program\\
GPU & Graphics Processing Unit\\
IoT & Internet of Things\\
IOU   & Intersect Over Union \\
JPG & Joint Photographic Experts Group\\
LSTM & Long Short-Term Memory\\
mAP   & mean Average Precision \\
MaskRCNN & Mask Region Convolutional Neural Network\\
PNG & Portable Network Graphics\\
R-CNN & Region-based Convolutional Neural Network (R-CNN) \\
ReLu & Rectified Linear Unit\\
ROI & Region Of Interest\\
SD & Secure Digital\\
Unicamp & State University of Campinas\\
SVM & Support Vector Machine \\
SWAP & Swap File\\
TP & True Positives\\
TN & True Negatives\\
UN & United Nations\\
USD & United States Dollars\\
YOLO & You Only Look Once\\
\end{tabular}
}

\bibliography{references}  






\end{document}